# Hybrid Approach to English-Hindi Name Entity Transliteration

Shruti Mathur
Department of Computer Engineering
Government Women Engineering College
Ajmer, India
shrutimathur19@gmail.com

Varun Prakash Saxena
Department of Computer Engineering
Government Women Engineering College
Ajmer, India
varunsaxena82@gmail.com

*Abstract*— Machine translation (MT) research in Indian languages is still in its infancy. Not much work has been done in proper transliteration of name entities in this domain. In this paper we address this issue. We have used English-Hindi language pair for our experiments and have used a hybrid approach. At first we have processed English words using a rule based approach which extracts individual phonemes from the words and then we have applied statistical approach which converts the English into its equivalent Hindi phoneme and in turn the corresponding Hindi word. Through this approach we have attained 83.40% accuracy.

## I. INTRODUCTION

Research in machine Translation (MT) is sixty years old. Still we do not have an MT engine which can provide a good translation. One of the problems for this is improper translation/transliteration of name entities (NEs). Most of the MT engines are not able to address this issue and thus provide poor quality translations. Name entities are basically nouns in a text which are categorized by predefined categories like person name, organization name, location, date and time etc. Most of the MT engines are unable to translate these texts properly. To understand this, let us consider some of the translations produced by Google MT Engine.

**English Sentence**: Ram is studying in Malviya National Institute of Technology.
**Hindi Translation**: राम प्रौद्योगिकी के मालवीय नेशनल इंस्टीट्यूट में पढ़ रहा है
**Observations:** "Ram" and "Malviya National Institute of Technology" are name entities. Only Ram was correctly transliterated. "Malviya National Institute of Technology" should have been transliterated as "मालवीय नेशनल इंस्टीट्यूट ऑफ टेक्नोलॉजी" or "मालवीय राष्ट्रीय प्रौद्योगिकी संस्थान".

**English Sentence**: Mary May is going to meet Susan June.
**Hindi Translation**: मैरी मई सुसान जून को पूरा करने जा रही है
**Observations:** "Mary May" and "Susan June" are names of people which are not transliterated properly.

**English Sentence**: Amber Fort was built by Raja Man Singh in 1592 AD in Jaipur.
**Hindi Translation**: एम्बर किले जयपुर में 1592 ई. में राजा आदमी सिंह ने बनवाया था.
**Observations:** "Amber", "Raja Man Singh", "1592 AD" and "Jaipur" are NEs. "Amber" and "Raja Man Singh" have not been translated/transliterated properly.

By looking at these examples we can clearly understand the need for proper mechanism to handle name entities as without it quality of translation would get affected. In this paper we shall be addressing this issue. We shall be devising a mechanism to implement a proper transliteration mechanism which will handle NEs and would help in improving the quality of machine translated output.

The rest of the paper is organized as follows: section II gives an outline of the work done to handle name entities and English-Hindi Machine Transliteration. Section II describes is approach. Section IV shows evaluation and results of the study and Section V concludes the paper.

## II. LITERATURE SURVERY

Many researchers have studied the use of machine transliteration for research in name entity translation or recognition. Babych and Hartley [1] implemented an automatic name entity recognition system which was done on the outputs generated by five different machine translation systems. They incorporated GATE's information extraction module in their systems concluded that combining IE technology with machine translation has a great potential for improving the overall output quality. Al-Onaizan and Knight [2] developed an algorithm for translation Arabic-English name entity phrases. They used both monolingual and bilingual resources and compared their results with the results produced by human translators and some commercial MT systems. They showed that their system had better correlation with human translators than any other system. They achieved an accuracy of 84%. Hassan et al. [3] performed a similar study which was done on extracted translation pairs. They showed that by using their approach the performance of a named entity translation system improves. Jiang et al. [4] used transliteration with web mining in translation of name entities. They used a maximum entropy based approach to train a classifier on pronunciation similarity, bilingual context and co-occurrence. This classifier was used to rank the candidate translations produced. Yeh et al. [5] proposed a pattern matching method for finding name entity's translation online. They developed an algorithm which automatically generated and weighted pattern which were used to search for name entities from bilingual corpus.

In an Indian context, Joshi and Mathur [6] proposed a phonetic mapping based algorithm for English-Hindi transliteration system which created a mapping table and a set of rules for transliteration of text. Joshi et al. [7] also proposed a predictive approach of for English-Hindi transliteration. Here instead of generated a single output they provided a list of possible text that can be selected by the user for correct transliteration. They looked at the partial text and tried to provide possible complete list as the suggestive list. Bhalla et al. [8] who used these two approaches for transliterating person and location name entities. Sharma et al. [9] trained a statistical machine translation system which could successfully translate English-Hindi name entities. They used Moses and Phrasal for this purpose. Moore [10] trained a classifier for English-Hindi transliteration using CRF based approach. They showed that using this approach we can successfully translate name entities with 85.79% accuracy and concluded that CRFs are best suited for processing Indian languages. Kharpa et al. [11] proposed a compositional machine transliteration where several transliteration approaches were combined to improve the accuracy. Their experiments showed the benefits of compositional methodology using some state of the art machine transliteration approaches. Agrawal and Singla [12] used three pronged approach in translating name entities. They used an aligner which generated English equivalents for Chinese name entities, a language model which improved the readability and a ranker which selected the best weighted translations. Ameta et al. [13] developed a transliteration system for Guajarati-Hindi language pair and used it in their Gujarati-Hindi translation engine which could effectively translate Gujarati name entities into Hindi. Bhalla et al. [14] used the Moses toolkit for generating translations for English-Punjabi name entities and claimed an accuracy of 88%.

### III. PROPOSED SYSTEM

#### A. Experimental Setup

In order to implement a transliteration system for English-Hindi, we first analyzed the spellings of different words and found that for Indian languages, most people use different spellings for the same words and all these spellings are taken to be correct. For example, let us consider the word "भारत".

For this Hindi word, different people would use the following spellings: Bharat, Bharath, Bhaaratha, and Bhaaraat. This is a non-exhaustive list and there can be many more variations to this word. Since this is a general phenomenon and cannot be captured by a rule based approach to transliteration. We tried to apply a hybrid approach to this. First we defined rules to capture phonemes of the English words. We identified that a word can be divided in seven different phonemes which are a group of vowels (V) and consonants (C). Table I shows these phonemes.

Once this was done, we collected the text from the web. We generally used news sites for this. We collected 10,000 sentences from these sites. In order to identify name entities we used Stanford's NER [15] tool for name entity extraction. In all we extracted 42,371 name entities. Table II shows the statistics of these name entities. After extraction of name entities, we applied our phonetic algorithm onto them and extracted different phonemes. Then each English phonemes was transliterated into Hindi, thus this created a knowledgebase of English and Hindi phonemes. We then used the ngram probability calculation [16] to generate the probabilities on this English-Hindi phoneme knowledgebase. We used equation 1 to compute the probabilities.

$$Prob(Hindi) = \frac{Count(English, Hindi)}{Count(English)} \quad (1)$$

Here, Prob(Hindi) was the probability of the Hindi phoneme, while Count(English, Hindi) was the count of the number of times a combination of a English and Hindi phonemes were seen in the knowledge base and Count(English) was the total count of the occurrence of a particular English phoneme in the knowledgebase. Table III shows the snapshot of this knowledge base.

TABLE I
PHONEMES WITH EXAMPLES

| S.No. | Combination |
|---|---|
| 1. | V |
| 2. | CV |
| 3. | VC |
| 4. | CVC |
| 5. | CCVC |
| 6. | CVCC |
| 7. | VCC |

TABLE II
STATISTICS OF NAME ENTITIES IN TRAINING CORPUS

| S.No. | Name Entity | Count |
|---|---|---|
| 1. | Person | 17,457 |
| 2. | Location | 11,548 |
| 3. | Organization | 8,569 |
| 4. | Date | 1,743 |
| 5. | Time | 1,656 |
| 6. | Misc | 1,398 |
| | **Total** | **42,371** |

TABLE III
SNAPSHOT OF ENGLISH-HINDI PROBABILITY KB

| English Phoneme | Hindi Phoneme | Probability |
|---|---|---|
| a | अ | 0.4532 |
| bh | भ | 0.3218 |
| i | ई | 0.4312 |
| ra | रा | 0.6532 |
| ro | रो | 0.4533 |
| shi | शि | 0.6788 |

*B. Methodology*

Since our major goal was to improve the accuracy in translating the name entities, we used the Stanford NER tool on English sentences and extracted the name entities. We used six class name entities as shown in table II. After extraction of English name entities, we applied our phonification algorithm and extracted the individual phonemes. Once this was done, we generated the equivalent Hindi phoneme for each English phoneme. Once all the phonemes for an English word were translated then we combined them to from an equivalent Hindi word. This is shown using the following algorithm and is shown in figure I.

**Input:** English Phoneme List
**Output:** Hindi Word
**Conversion Algorithm**
1. Input the English Phoneme List as phoneme.
2. Read English-Hindi Probability KB as KB
3. phlen = phoneme.length
4. count = 1
5. repeat steps 5 to 8 till count <= phlen
6. generate list of English-Hindi phonemes for phoneme[count] with their respectively probability
7. hinpho[count] = max_prob(phoneme[count])
8. count += 1
9. combine hinpho to form Hindi word as hword
10. return hword

The following examples explain the working of the entire system:

**English Sentence:** Ram is going to Bhopal
**Stanford NER Output:** Ram/Person; Bhopal/Location
**Phonification Module:**    Ram => Ra m
                              Bhopal => Bho pa l
**Conversion Module:**    Ra => रा
                          m => म        राम
                          Bho => भो
                          pa => पा
                          l => ल        भोपाल

In this sentences Ram and Bhopal are name entities and are identified by NER module. These name entities are then passed onto the phonification module which separately generates phonemes for each word. Ram is split into two phonemes and Bhopal is split into three phonemes. This is then passed onto the conversion module which then transforms these English phonemes into Hindi phonemes and in turn combines them to form the complete word as राम and भोपाल.

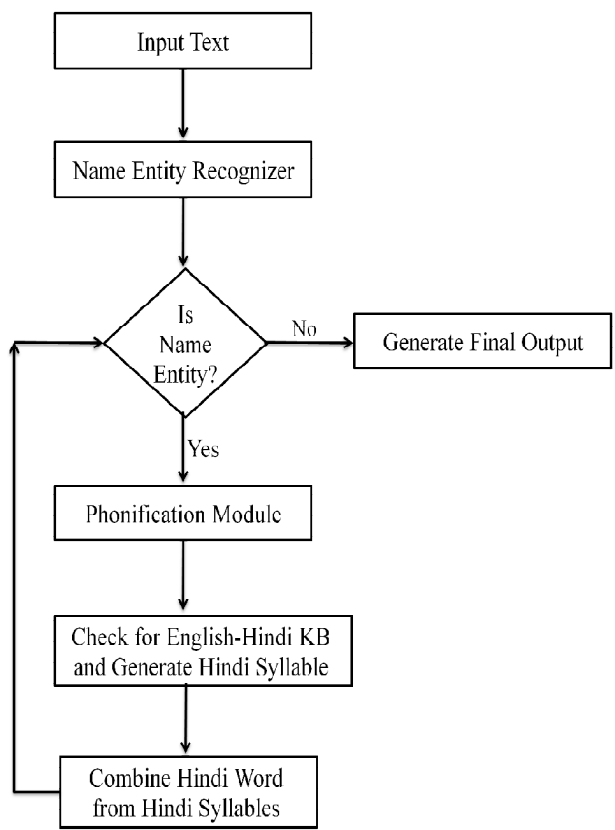

Fig. I English-Hindi Name Entity Transliteration System

**English Sentence:** Ramesh is going to Delhi to meet Suresh.
**NER Module:**    Ramesh/Person; Delhi/Location; Suresh/Person
**Phonification Module:**    Ramesh => Ra me sh

                              Delhi => De lhi

                              Suresh => Su re sh
**Conversion Module:**    Ra => रा
                          me => मे
                          sh => श        रमेश
                          De => दे
                          lhi => लही     देलही

Su => सु
re => से
sh => श    सुरेश

In this sentences Ramesh, Suresh and Delhi are name entities. These name entities are then passed onto the phonification module which generates their phonemes. Ramesh is split into three phonemes, Suresh is also split into three phonemes and Delhi is split into two phonemes. This is then passed onto the conversion module which then transforms these English phonemes into Hindi phonemes and in turn combines them to form the complete word as रमेश, सुरेश and देलही.

IV. EVALUATION

Our basic objective was to develop a mechanism which can effectively handle name entities. This would act a sub-module in any machine translation system and in turn would improve the quality of translation. But in order to incorporate this into any system, we first need to test and verify its accuracy. Thus in order to check the accuracy of the system, we collected 1000 sentences which were not part of the training corpus. These sentences had 9234 name entities. The statistics of this corpus is shown in table IV

TABLE IV
STATISTICS OF NAME ENTITIES IN TEST CORPUS

| S.No. | Name Entity | Count |
|---|---|---|
| 1. | Person | 5,263 |
| 2. | Location | 2,770 |
| 3. | Organization | 1,108 |
| 4. | Date | 13 |
| 5. | Time | 27 |
| 6. | Misc | 53 |
|  | Total | 9,234 |

We applied this test corpus onto our system and calculated the accuracy according to standard precision, recall and f-measure. This was done by comparing the results of generated by a human translator who manually transliterated these name entities. They are calculated as per the following equation:

$$Precision\ (P) = \frac{correct}{System\ Output} \quad (2)$$

$$Recall\ (R) = \frac{correct}{Reference\ Output} \quad (3)$$

$$F - Measure = \frac{2 \times P \times R}{P+R} \quad (4)$$

Here, system output is the no. of name entities generated by our system. Reference output is the no. of name entities generated by the human translator and correct is the no. of correct matches between our system's and the human translator are output.

In order to provide a level playing field, we gave the output of the NER module to the human translator as well. Table V shows the results of this evaluation.

TABLE V
Summary of Evaluation

| Total Name Entities | 9,234 |
|---|---|
| System Generated Name Entities | 9,180 |
| Human Generated Name Entities | 9,234 |
| Correct Name Entities | 7,679 |
| Precision | 0.8365 |
| Recall | 0.8316 |
| F-Measure | 0.8340 |

We attained an accuracy of 83.40%. The reason for this low score was that, our system could very well transliterate the name entities of type Person, Location, Date and Time but most of the name entities of type Organization are not transliterated, they are actually translated. For example, National Institute of Technology is translated into राष्ट्रीय प्रौद्योगिकी संस्थान whereas our system gave an output as नेशनल इंस्टिट्यूट ऑफ टेक्नोलोजी. According to human translator's reference, this was considered as wrong output. Table VI shows entity wise outputs of our system.

TABLE VI
ENTITY WISE ANALYSIS OF EVALUATION

| S.No. | Name Entity | Count | System Output | Correct |
|---|---|---|---|---|
| 1. | Person | 5,263 | 5,263 | 4,893 |
| 2. | Location | 2,770 | 2,770 | 2,603 |
| 3. | Organization | 1,108 | 1,107 | 143 |
| 4. | Date | 13 | 13 | 13 |
| 5. | Time | 27 | 27 | 27 |
| 6. | Misc | 53 | 0 | 0 |
|  | Total | 9,234 | 9,180 | 7,679 |

V. CONCLUSION

In this paper we have shown the implementation of a Name Entity Transliteration system for English-Hindi language pair. The system was developed on a hybrid model where English name entities were processing using a set of rules and the conversion of English-Hindi was done statistically. The system did fairly well with all name entities, except for Organization. This was because most organization names change were they are converted from English to Hindi. Thus an immediate future enhancement of this system is to incorporate a translation mechanism to handle this kind of name entities and improve the accuracy of the system. Moreover, some more rules are needed to be added to phonification module, so that better transliterations can be produced.